\documentclass{article}

\usepackage{arxiv}

\usepackage[utf8]{inputenc} 
\usepackage[T1]{fontenc}    
\usepackage{hyperref}       
\usepackage{url}            
\usepackage{booktabs}       
\usepackage{amsfonts}       
\usepackage{nicefrac}       
\usepackage{microtype}      
\usepackage{lipsum}
\usepackage{graphicx}
\usepackage{subfig}
\usepackage{amsmath}
\usepackage{algorithm}
\usepackage{algpseudocode}

\title{NEAT and HyperNEAT based Design for Soft Actuator Controllers}

\author{
    Hugo Alcaraz-Herrera\\
    Unconventional Computing Laboratory, \\
    College of Arts, Technology and Environment, \\
    University of the West of England, \\
    Bristol, BS16 1QY, United Kingdom \\ \texttt{hugo.alcaraz@uwe.ac.uk}
\And
    Michail-Antisthenis Tsompanas  \\
    Unconventional Computing Laboratory \& \\
    School of Computing \& Creative Technologies,\\
    College of Arts, Technology and Environment, \\
    University of the West of England,\\ 
    Bristol, BS16 1QY, United Kingdom \\
    \texttt{antisthenis.tsompanas@uwe.ac.uk}
\And
       Igor Balaz\\
       Laboratory for Meteorology, Physics and Biophysics,\\ Faculty of Agriculture, \\
       University of Novi Sad, \\ Trg Dositeja Obradovica 8, 21000, Novi Sad, Serbia
\And
       Andrew Adamatzky \\
       Unconventional Computing Laboratory,\\
    College of Arts, Technology and Environment, \\
    University of the West of England,\\ 
    Bristol, BS16 1QY, United Kingdom \\
}

\begin{document}
\maketitle

\begin{abstract}
Since soft robotics are composed of compliant materials, they perform better than conventional rigid robotics in specific fields, such as medical applications. However, the field of soft robotics is fairly new, and the design process of their morphology and their controller strategies has not yet been thoroughly studied. Consequently, here, an automated design method for the controller of soft actuators based on Neuroevolution is proposed. Specifically, the suggested techniques employ Neuroevolution of Augmenting Topologies (NEAT) and Hypercube-based NEAT (HyperNEAT) to generate the synchronization profile of the components of a simulated soft actuator by employing Compositional Pattern Producing Networks (CPPNs). As a baseline methodology, a Standard Genetic Algorithm (SGA) was used. Moreover, to test the robustness of the proposed methodologies, both high- and low-performing morphologies of soft actuators were utilized as testbeds. Moreover, the use of an affluent and a more limited set of activation functions for the Neuroevolution targets was tested throughout the experiments. The results support the hypothesis that Neuroevolution based methodologies are more appropriate for designing controllers that align with both types of morphologies. In specific, NEAT performed better for all different scenarios tested and produced more simplistic networks that are easier to implement in real life applications.
\end{abstract}


\section{Introduction}\label{sec:introduction}

A subset of robotics research is focused on soft robots, namely machines that are developed from flexible and adaptable materials \cite{Rus2015}. These robots surpass conventional rigid ones in specific applications with special requirements, such as healthcare \cite{Hsiao2019} or search and rescue \cite{Milana2022}. Also, most of the prototypes build so far are inspired by the movement and behaviour of living creatures \cite{park2016phototactic}.

Soft robots offer great potential for various applications, but they also come with significant challenges. One major difficulty lies in designing an effective structure, as their flexible materials exhibit complex mechanical properties that are difficult to define and predict \cite{Hiller2014}. Conventional robotic design relies on testing multiple physical prototypes through an iterative process that demands substantial time and material resources \cite{Schulz2016}, whereas, the design process for soft robots is even more complicated due to the non-linear behavior of their materials. This makes accurate modeling and effective optimizing their morphology and control strategy particularly challenging.

Controlling soft robots requires addressing two key challenges. First, their flexible materials can be distorted by actions of the robot itself or the environment at any point, leading to countless degrees of freedom. Second, these materials exhibit non-linear and time-dependent properties, making their behavior unpredictable over time. Due to these factors, accurately modeling and controlling the movement of soft robots is a complex task \cite{Wang2022}. On the other hand, these countless degrees of freedom can accommodate the adoption of elaborate behaviour and efficient movement through embodied intelligence principles \cite{gupta2021embodied}.

A promising approach for designing controllers for soft robots is Neuroevolution (NE), which involves using genetic algorithms (GAs) to evolve the structure and weights of artificial neural networks (ANNs). Among NE methods, Neuroevolution of Augmenting Topologies (NEAT) has demonstrated high efficiency \cite{Stanley2002}. Building on the idea that natural structures often feature repeating patterns, an advanced extension of NEAT, called Hypercube-based Neuroevolution of Augmenting Topologies (HyperNEAT), was proposed \cite{Stanley2009}. HyperNEAT employs a specialized type of neural networks known as Compositional Pattern-Producing Networks (CPPNs), which differ from traditional ANNs by utilizing a variety of mathematical functions (e.g., sine and square waves) to generate patterns such as symmetry and repetition. This ability to create structured patterns enables the evolution of more complex and effective network topologies \cite{Stanley2007cppn}.

This research builds upon previous work \cite{Alcaraz2024controllers} to determine the appropriateness of NEAT and HyperNEAT as design methods for controllers of {\em soft actuator morphologies} (SAMs). In specific, the ability of NEAT and HyperNEAT in designing controllers for diverse types of SAMs was investigated. Two types of SAMs were investigated; namely, the best performing individuals after evolutionary optimization process performed in a previous work \cite{Alcaraz2024actuator} and the least good performing individuals at the end of the same evolutionary process. The fitness of the individuals is considered to be the longest upward bending movement to a given SAM. That SAM may be embodied into a catheter for targeted drug delivery to areas of the human body that are difficult to reach, and that application is considered when setting the parameters and constraints of the simulator that provide the fitness function for SAMs \cite{Tsompanas2024}. Moreover, the robustness of the best evolved controllers was tested by applying them to morphologies with which they were not optimized. Finally, we investigated the feasibility of evolved controllers, manifested as neural networks, by evaluating the number of nodes and connections of these networks and the activation functions usage frequency.

The rest of the paper is organised as follows. Section~\ref{sec:background} briefly outlines background research related to NEAT and HyperNEAT and their usage as controller design engines. Section~\ref{sec:neuroevolution} introduces the mechanism of NEAT and HyperNEAT. Section~\ref{sec:experimental_setup} presents the experimental setup utilised to analyse the effectiveness of the NE-based techniques. Section~\ref{sec:experiments_comparison} describes the results provided by the experiments performed in the task of designing controllers under NEAT and HyperNEAT. Finally, Section~\ref{sec:conclusions} concludes this work, presents the insights gained during experimentation and outlines future work.


\section{Background}\label{sec:background}

NEAT has been applied in robotic control design, particularly in navigating complex environments. For example, it has been used to develop controllers for robots operating in crowded spaces \cite{Seriani2021}. In industrial robotics, a ray-casting model can serve as the robot's perception system, allowing it to detect objects in real time while retaining memory of previously perceived obstacles. Initially, controllers are tested within simulated environments before being implemented in a simplified physical setup. Findings indicate that NEAT-generated controllers successfully adapt and converge to effective designs in both simulated and real-world scenarios.

Another study explores the application of NEAT in designing locomotion controllers \cite{Tibermacine2014}. The research investigates virtual creatures simulated within the {\em Open Dynamic Engine} (ODE) physics engine. NEAT-generated controllers are compared to those developed using a more conventional evolutionary approach \cite{Ruebsamen2002}. The evaluation is based on three locomotion tasks: crawling, running, and performing a somersault. The results suggest that NEAT-designed controllers outperform traditional methods, as they better account for the creatures' morphological features and demonstrate greater robustness.

Furthermore, NEAT was used to generate controllers to enhance the locomotion of a snake-like modular robot \cite{Song2023}. In order to analyse the performance of the controllers designed by NEAT, a Multi-objective Genetic Algorithm (MOGA) is employed as a baseline under three different environments: (i) the snake-like robot is surrounded by cylindrical obstacles; (ii) stairs surround the snake-like robot; and (iii) the snake-like robot is under a low ceiling. Results showed that NEAT outperformed MOGA in all the environments used for experimentation.

Under the same research path, an approach based on NEAT focused on designing controllers to enable a physical robotic system to grasp unforeseen objects in an unstructured environment \cite{Huang2014}. Experiments consisted of predicting the suitable hand configuration in terms of positions and orientations for grasping a target object. After simulations, the controller was transferred to a physical robot called {\em Dreamer} robot. Results point out that NEAT can enhance the performance of grasping tasks when there is no previous knowledge of objects to grasp or the environment.

NEAT has also been applied in the development of controllers for autonomous vehicles. One study explored a reactive navigation hybrid controller for non-holonomic mobile robots \cite{Caceres2017}. The experiments were carried out using a custom simulation platform based on the kinematic model of a vehicular robot. The findings indicate that NEAT-designed controllers successfully managed the robot’s kinematics in an unfamiliar environment, effectively avoiding obstacles and reaching the target destination.

HyperNEAT has also been employed for designing both robot controllers and their physical structures. A notable study introduced a method for optimizing robot morphologies alongside their controllers \cite{Tanaka2022}. This approach was evaluated across four scenarios, assessing the adaptability of the robots in different conditions. The findings indicated that the method effectively generated viable morphologies and produced controllers capable of executing tasks successfully.

HyperNEAT has also been applied to develop locomotion controllers for autonomous crawler robots equipped with flippers \cite{Sokolov2017}. To assess their effectiveness, the controllers were evaluated using the ROS/Gazebo simulation platform. The CPPNs received inputs such as Light Detection and Ranging data, the robot's position and orientation, flipper angles, and track velocities; while the outputs generated movement commands for the flippers and tracks. The results demonstrated that HyperNEAT-based controllers effectively enabled crawler robots to navigate complex three-dimensional environments and escape obstructions.

Following the same research path, HyperNEAT has designed homogeneous yet specialised controllers for modules that compose multi-robot organisms \cite{Haasdijk2010}. The approach was assessed in the task of developing reactive four-legged gaits. The results pointed out that controllers showed autonomous behaviour, and they could allow the successful locomotion of a robot by the local exchange of information.

HyperNEAT has also been used to design controllers for driving purposes. A study presents a simulation of autonomous robots whose controller was designed by HyperNEAT \cite{Drchal2009}. The robots employed a $180^\circ$ wide sensor array, which simulated a camera as input for the controller that could be used in a real robot. The simulations of robots were performed on ViVAE (Visual Vector Agent Environment), and they were trained to drive, maximising the average speed and trying to avoid collisions with obstacles and with other robots. Results indicated that HyperNEAT was able to design controllers capable of driving at an acceptable speed and avoiding collisions in a limited number of evolutionary steps.

A study proposed the use of HyperNEAT to design controllers for quadruped robots \cite{Risi2013}. The approach took robot morphologies as input and produced neural-network-based controllers adaptable to different structures. The controllers' performance was evaluated across three distinct morphologies and compared to static controllers. The findings indicated that HyperNEAT effectively identified the relationship between morphology and controller architecture, leading to improved adaptability and performance.

Inspired by the aforementioned works of employing NE for controlling different types of robots (i.e., crawlers, runners and even car-like) to achieve higher displacement, the same methodology is tested here on robots limited to angular-only movement. The insights gained by this research will help to design robotic controllers' optimization methodologies for tasks requiring precision movements, such as medical applications.


\section{Neuroevolution}\label{sec:neuroevolution}

This section describes in detail NEAT \cite{Stanley2002} and HyperNEAT \cite{Stanley2009}, two well-known NE-based algorithms that are employed in this research.

\subsection{NEAT}\label{sec:neuroevolution_neat}

NEAT is an algorithm designed to counteract three primary issues identified in previous NE-based algorithms \cite{Stanley2002}: (i) the lack of suitable solution representations capable of recombining neural networks with arbitrary topologies; (ii) avoiding the premature disappearance of novel topological structures discovered throughout the evolutionary process; and (iii) circumventing the use of fitness functions that punish complex topologies of networks (i.e., individuals).

\subsubsection{Historical markings to track genes.}\label{sec:neuroevolution_neat_genes}

NEAT counteracts the first issue as the representation of networks consists of a list of connection genes encoding a connection between two nodes. Each gene contains the ``origin'' and ``destination'' nodes, the weight of the connection, the state of the connection (enabled or disabled), and one {\em global innovation number}, which is a unique numerical ID utilised to perform the crossover operator.

\begin{figure}[tb!]
   \centering
   \includegraphics[width=1.0\linewidth]{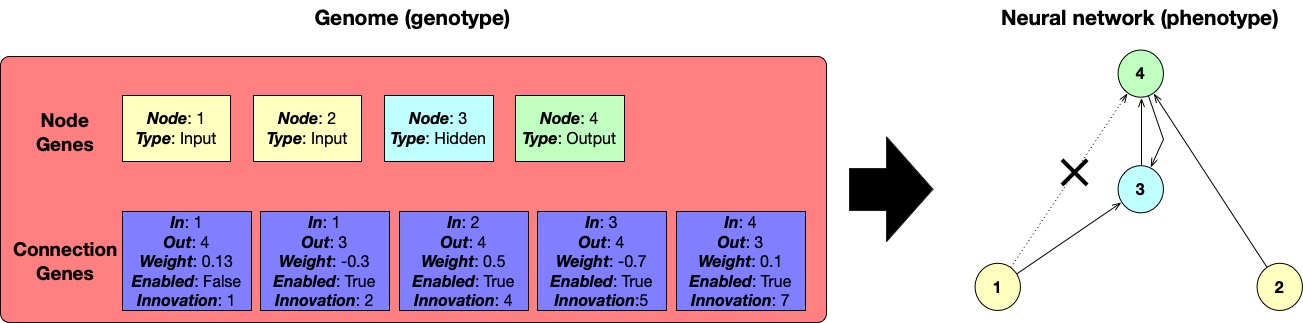}
   \caption{Example of a genotype and its phenotype under NEAT. Figure adapted from \cite{Stanley2002efficient}.}
   \label{fig:neuroevolution_neat_genotype}
\end{figure}

Two genes with the same global innovation number represent the same structure. Consequently, the genes of both genomes having the same innovation number are aligned, and the rest of the genes are inherited from the fittest parent or randomly chosen. When the mutation operator generates a new gene, the innovation number is incremented and associated with the new gene. In general, innovation number can be seen as the chronological history of all genes during evolution. An example of a genotype (i.e., a neural network) and its phenotype is depicted in Fig.~\ref{fig:neuroevolution_neat_genotype}.

\subsubsection{Protecting innovation by speciation.}\label{sec:neuroevolution_neat_speciation}

The population can be split into species using topological similarities since NEAT employs historical markings (as mentioned in the previous section). Therefore, individuals belonging to one of the species interact with their associated peers. The canonical method to measure the compatibility between two individuals considers the number of excess and disjoint genes. Thus, the more disjoint genes are, the less compatible two individuals (i.e., neural networks) are. Under this scheme, the compatibility distance ($\delta$) between two genomes is calculated as follows:

\begin{equation}\label{eq:neuroevolution_neat_distance} 
     \delta = \frac{c_{1}E}{N} + \frac{c_{2}D}{N} + c_3 \times \overline{W}
\end{equation}

\noindent
where $E$ represents the number of excess genes, $D$ represents the disjoint genes, and $\overline{W}$ is the mean weight differences of matching genes. With respect to $c_1$, $c_2$, and $c_3$, they are utilised to weight the relevance of $E$, $D$, and $\overline{W}$, respectively. Furthermore, $N$ represents the number of genes in the individual with the largest number of genes; this is used to normalise the genome size. The distance $\delta$ assigns individuals to species by a compatibility threshold $\delta_t$. For each individual, if the distance to a randomly chosen member of the species is $<\delta_t$, the individual is assigned to this species where the condition is met, assuring that the individual is not assigned to more than one species simultaneously.

NEAT employs {\em explicit fitness sharing} to reproduce individuals. Under this mechanism, species' members share their fitness, which prevents any population from taking over the population \cite{Goldberg1987}. In general, this approach adjusts the fitness of individuals by dividing the original fitness by the number of individuals in each species. Consequently, species grow or shrink if their average adjusted fitness is above or below the population average. This mechanism aims to alleviate the second issue.

\subsubsection{Minimal topologies.}\label{sec:neuroevolution_neat_minimal}

The novelty protection mechanism of NEAT (i.e., historical markings) also allows for the initialisation of individuals minimally: without any hidden nodes and with the input neurons being fully connected to output neurons. When the initial individuals start evolving, new structures are introduced incrementally to topologies, and those that achieve a suitable performance survive.

The initialisation procedure of NEAT offers a searching behaviour that considers a minimal number of weight dimensions first. This leads to a significant reduction in computational cost for finding solutions and avoids evaluating complex structures that are not crucial for the evolutionary process. In this way, the third issue is alleviated.

\subsection{HyperNEAT}\label{sec:neuroevolution_hyperneat}

Arguably, the most popular extension of NEAT is HyperNEAT. This algorithm uses NEAT to evolve the topology of {\em Compositional Pattern-Producing Networks} (CPPNs) \cite{Stanley2007cppn}, a specific type of ANNs since they can reproduce patterns observed in Nature, such as symmetry, repetition, and regularity \cite{Stanley2009}. The difference between NEAT and HyperNEAT consists of two aspects: (a) Substrate and (b) Activation functions.

\subsubsection{Substrate.}\label{sec:neuroevolution_hyperneat_substrate}

HyperNEAT can embody the geometry observed of the domain problem using the pattern reproduction features of CPPNs. Therefore, it considers those geometrical aspects to design the topology of a neural network called {\em substrate}.

\begin{figure}[tb!]
   \centering
   \includegraphics[width=0.51\linewidth]{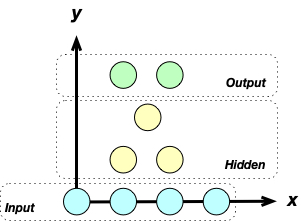}
   \caption{An example of a two-dimensional substrate configuration known as {\em grid}.}
   \label{fig:neuroevolution_hyperneat_substrate}
\end{figure}

Substrates have numerous configurations. Arguably, the most well-established substrate configuration is the {\em grid}, a set of nodes allocated in a two-dimensional plane. Another efficient substrate configuration is the {\em three-dimensional grid}, which is defined by a set of nodes allocated in a three-dimensional space. Figure~\ref{fig:neuroevolution_hyperneat_substrate} depicts an example of a two-dimensional substrate or grid.

Since CPPNs are used to compute the topology of ANNs embodied in the substrate utilising the position of neurons, they are queried depending on the substrate configuration. Thus, when a two-dimensional substrate is used, the weight between the neuron $m$ and the neuron $h$ is obtained utilising their position in the grid:

\begin{equation}\label{eq:neuroevolution_hyperneat_2d_substrate_weight} 
     CPPN(x_h,y_h,x_m,y_m) = weight_{hm}
\end{equation}

\noindent
In order to calculate the bias of neurons in the substrate, CPPNs are queried using the coordinates of the ``origin'' neuron, whereas the coordinates of the ``destination'' position are set to zero. Thus, the bias of the neuron $m$ is obtained as follows:

\begin{equation}\label{eq:neuroevolution_hyperneat_2d_substrate_bias} 
    CPPN(x_m,y_m,0,0) = bias_m
\end{equation}

\noindent
When a three-dimensional grid is being used, the weight between the neuron $h$ and the neuron $m$ is calculated by:

\begin{equation}\label{eq:neuroevolution_hyperneat_3d_substrate_weight} 
    CPPN(x_m,y_m,z_m,x_h,y_h,z_h) = weight_{mh}
\end{equation}

\noindent
Furthermore, the bias of neuron $m$ is obtained as follows:

\begin{equation}\label{eq:neuroevolution_hyperneat_3d_substrate_bias} 
    CPPN(x_m,y_m,z_m,0,0,0) = bias_m
\end{equation}

\noindent
Typically, CPPNs' outputs are in the $[-1.0,1.0]$ range, which is essential to consider if a normalisation procedure is required. In general, the objective of HyperNEAT consists of evolving the topology and weights of ANNs, which are represented by a substrate with a geometry that is determined by employing CPPNs. Furthermore, Algorithm~\ref{alg:neuroevolution_hyperneat} depicts the general framework of HyperNEAT.

\begin{algorithm}
  \caption{General framework of HyperNEAT.}
    \label{alg:neuroevolution_hyperneat}
\begin{algorithmic}
  \State  initialise~substrate
  \State  initialise~population~of~CPPNs
  \While{termination condition not met}
     \For{each CPPN}
         \State  generate~connections~for~substrate
         \State  generate~bias~for~substrate
         \State  assign~aptitude~to~CPPN
     \EndFor
     \State  evolve CPPNs through NEAT
  \EndWhile
\end{algorithmic}
\end{algorithm}

\subsubsection{Activation functions}\label{sec:neuroevolution_hyperneat_activation}

Usually, NEAT-based designs use sigmoid function as an activation function in hidden nodes. On the other hand, HyperNEAT can employ numerous activation functions composing CPPNs, such as Gaussian, trigonometric, and periodic. The use of different activation functions, when CPPNs are evolving, allows a significantly wider exploration of the search space of network topologies.


\section{Experimental setup}\label{sec:experimental_setup}

In order to assess the suitability of NEAT and HyperNEAT as controller design engines for SAMs, three elements must be considered: (i) the physics engine implemented to simulate SAMs; (ii) the specific configuration utilised for NEAT and HyperNEAT for experimentation; and (iii) the evolutionary setup employed during experimentation.

\subsection{Voxelyze}\label{sec:experimental_setup_voxelyze}

Due to the dynamics of SAMs, 
a three-dimensional space $(x,y,z)$ is used as a design canvas. Moreover, to represent a case study for a drug-delivering catheter empowered by the proposed SAM, the insertion point of the catheter, i.e., one end of the SAMs is considered fixed during simulation.

A physics engine called {\em Voxelyze} \cite{KriegmanGitHub} is employed to simulate the mechanical response of SAMs considering specific conditions of the environment, such as viscosity and gravity. A {\em voxel}, the minimum building block in the Voxelyze context, can represent diverse types of material. Hence, stacking voxels in three dimensions assembles a SAM. In the scope of this research and for simplicity in relation to the design landscape, two types of voxels are implemented: one active and one passive, representing the mechanical properties of contractile and immotile building blocks, respectively.

Based on previous research in which {\em in vitro} experiments were conducted \cite{Kriegman2020}, the parameters of the mechanical properties considered during simulation are: Poisson's ratio $0.35$, Young's modulus $5\times10^6$ Pa, coefficients of static and dynamic friction set to $1.0$ and $0.5$, respectively. Furthermore, active voxels were characterised by volumetric actuation of $\pm 50\%$ of the volume at rest during $4$ Hz cycles.

In this research, Voxelyze is considered in order to provide the fitness function of controllers, i.e., induce the longest upward bending movement to specific SAMs. Therefore, the physics engine was modified to trace the position of SAMs in the $x,y,z$ axes during a determined time frame \cite{TsompanasGitHub}. In addition, SAMs were designed inside a passive enclosure to follow and be comparable with results from previous research \cite{Tsompanas2024}. This design decision represents a bioreactor with a nutrient supply for a muscle actuator.

The output of Voxelyze consists of the trace of the free end of the simulated SAM in space starting from the initial position ($t=0$) and ending at the final position ($t=n$). Also, the number of voxels composing the SAM are reported. In terms of maximum voxel design space, the dimensions of SAMs are 20 units on the $x$ axis and 8 units on the $y$ and $z$ axes.

\subsection{NEAT and HyperNEAT configuration}\label{sec:experimental_setup_configuration}

Although NEAT and HyperNEAT use CPPNs as the primary element of their mechanism, these algorithms use them in a different manner. Each configuration is described in the following sections.

\subsubsection{NEAT configuration.}\label{sec:experimental_setup_configuration_neat}

In this research, an extended version of NEAT, known as {\em CPPN-NEAT}, is implemented \cite{Stanley2007cppn}. Since a discrete three-dimensional canvas is used to simulate SAMs, and two types of voxels representing different materials, the input of controllers, which are encoded as CPPNs, contains: (i) the coordinates of each point, and therefore the voxel, across the canvas and (ii) the type of material. Furthermore, the output of controllers is the phase offset of each voxel across the canvas, which determines the delay in the expansion behaviour of active voxels. Equation~\ref{eq:experimental_setup_configuration_neat}, describes how CPPNs are queried.

\begin{equation}\label{eq:experimental_setup_configuration_neat} 
    CPPN(x_i,y_i,z_i,m_i) = pho_i
\end{equation}

\noindent
where $(x_i,y_i,z_i)$ tuple is the coordinates of the $i$-th point in the three-dimensional canvas. Regarding $m_i$, it represents the material type of the voxel located in the $i$-th point in the canvas, which is encoded as follows: $0$, absence of a voxel; $1$, passive voxel; $3$ contractile voxel. Finally, $pho_i$ is the phase offset of the voxel located in the $i$-th point of the three-dimensional canvas. It is important to highlight that the output of CPPNs (i.e., controllers) is clamped in the $[-2\pi,2\pi]$ range to provide a complete sinusoidal-like contraction per voxel.

\subsubsection{HyperNEAT configuration.}\label{sec:experimental_setup_configuration_hyperneat}

When HyperNEAT is applied, the substrate is first defined (see Section~\ref{sec:neuroevolution_hyperneat_substrate}). Since a three-dimensional canvas is used to simulate SAMs, and the material type is encoded with one number, the substrate has four input neurons. Moreover, one output neuron is considered to provide the phase offset of each voxel throughout the canvas. Thus, substrates are queried as follows:

\begin{equation}\label{eq:experimental_setup_configuration_hyperneat} 
    substrate(x_i,y_i,z_i,m_i) = pho_i
\end{equation}

\noindent
where $(x_i,y_i,z_i)$ tuple represents the coordinates of the $i$-th point in the three-dimensional canvas, and $m_i$ is the material type of voxels, whose values are encoded in the same manner as NEAT (see Equation~\ref{eq:experimental_setup_configuration_neat}). Regarding $pho_i$, it represents the phase offset of the voxel located in the $i$-th point in the three-dimensional canvas.

\begin{figure}[tb!]
  \centering
     \includegraphics[width=0.51\linewidth]{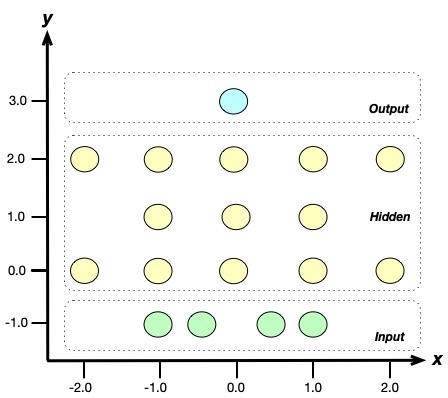}
  \caption{Substrate utilised under HyperNEAT to design SAM controllers. Image taken from \cite{Alcaraz2024controllers}.}
  \label{fig:experimental_setup_configuration_hyperneat_substrate}
\end{figure}

Once the input and output neurons of the substrate have been defined, the next step focuses on determining the position of the neurons composing the substrate, which has been determined by a series of experiments where the number of layers and neurons per layer is varied in the $[3, 10]$ range to discover the optimal neuron allocation. Figure~\ref{fig:experimental_setup_configuration_hyperneat_substrate} shows the design of the two-dimensional substrate implemented for experimentation in this research.

Due to the fact that the substrate has been designed in a two-dimensional space, CPPNs are queried using Equation~\ref{eq:neuroevolution_hyperneat_2d_substrate_weight} to compute the weights among the neurons allocated in the substrate, whereas Equation~\ref{eq:neuroevolution_hyperneat_2d_substrate_bias} is utilised to calculate the bias of neurons composing the substrate. In addition, based on the definition of HyperNEAT \cite{HyperNEAT2015}, a threshold is required to determine the presence of the connection between neuron $h$ and neuron $m$. Thus, if $|weight_{hm}| < 0.2$, there is no connection between neurons $h$ and $m$. Otherwise $weight_{hm}$ is normalised in the $[-3.0, 3.0]$ range.

Finally, the activation function utilised for the neurons of the substrate is ReLU since it presents representational sparsity properties and induces a linear (or close to linear) behaviour \cite{Glorot2011}.

\subsection{Evolutionary configuration}\label{sec:experimental_evolutionary_configuration}

Following the configuration used in previous studies \cite{Alcaraz2024controllers}, for NEAT and HyperNEAT 50 individuals (i.e., CPPNs) comprise the population. Furthermore, each evolutionary run lasts 200 generations. In addition, individuals are initialised without hidden neurons, whereas input neurons are fully connected to output neurons (i.e., minimal topology). The specific parameters for evolving CPPNs are shown in Table~\ref{tab:setup_neat_hyperneat}.

\begin{table}
\caption{Parameters utilised to evolve CPPNs under NEAT and HyperNEAT. Table taken from \cite{Alcaraz2024controllers}.}
\label{tab:setup_neat_hyperneat}
 \begin{center}
  \begin{tabular}{ c c } 
   \toprule
   Parameter & Value \\
   \midrule
   compatibility threshold & 3 \\
   compatibility disjoint coefficient & 1.0 \\ 
   compatibility weight coefficient & 0.5 \\
   maximum stagnation & 25 \\
   survival threshold & 0.2 \\ 
   activation function mutate rate & 0.4 \\
   adding/deleting connection rate & 0.2/0.1 \\
   activating/deactivating connection rate & 0.5\\
   adding/deleting node rate & 0.2/0.1 \\
   \bottomrule
  \end{tabular}
 \end{center}
\end{table}

In order to analyse the effect of activation functions in the performance of NE-based approaches, two different dictionaries of activation functions are implemented. The full dictionary (FD) contains the following activation functions: {\em Sine}, {\em Negative sine}, {\em Absolute value}, {\em Negative absolute value}, {\em Squared}, {\em Negative squared}, {\em Squared absolute value}, {\em Negative squared absolute value}, {\em Sigmoid}, {\em Clamped}, {\em Cubical}, {\em Exponential}, {\em Gaussian}, {\em Hat}, {\em Identity}, {\em Inverse}, {\em Logarithmic}, {\em ReLU}, {\em SeLU}, {\em LeLU}, {\em eLU}, {\em Softplus}, {\em Hyperbolic tangent}. The reduced dictionary (RD), is a subset of FD and is composed of:  {\em Sine}, {\em Negative sine}, {\em Squared}, {\em Negative squared}, {\em Sigmoid}, {\em Cubical}, {\em Gaussian}, {\em Logarithmic}, {\em Hyperbolic tangent}. Although the FD was used in previous works \cite{Alcaraz2024locomotion,Alcaraz2024actuator} to avoid the restriction of realizing certain abrupt patterns, the comparison with a RD (which includes a subset of the FD with functions that tend to produce a more gradual and soft variance in space) is demonstrated here to reveal whether the extra activation functions are actually necessary or induce an extra overhead in the search process. 

\begin{figure}[tb!]
  \centering
     \includegraphics[width=.73\linewidth]{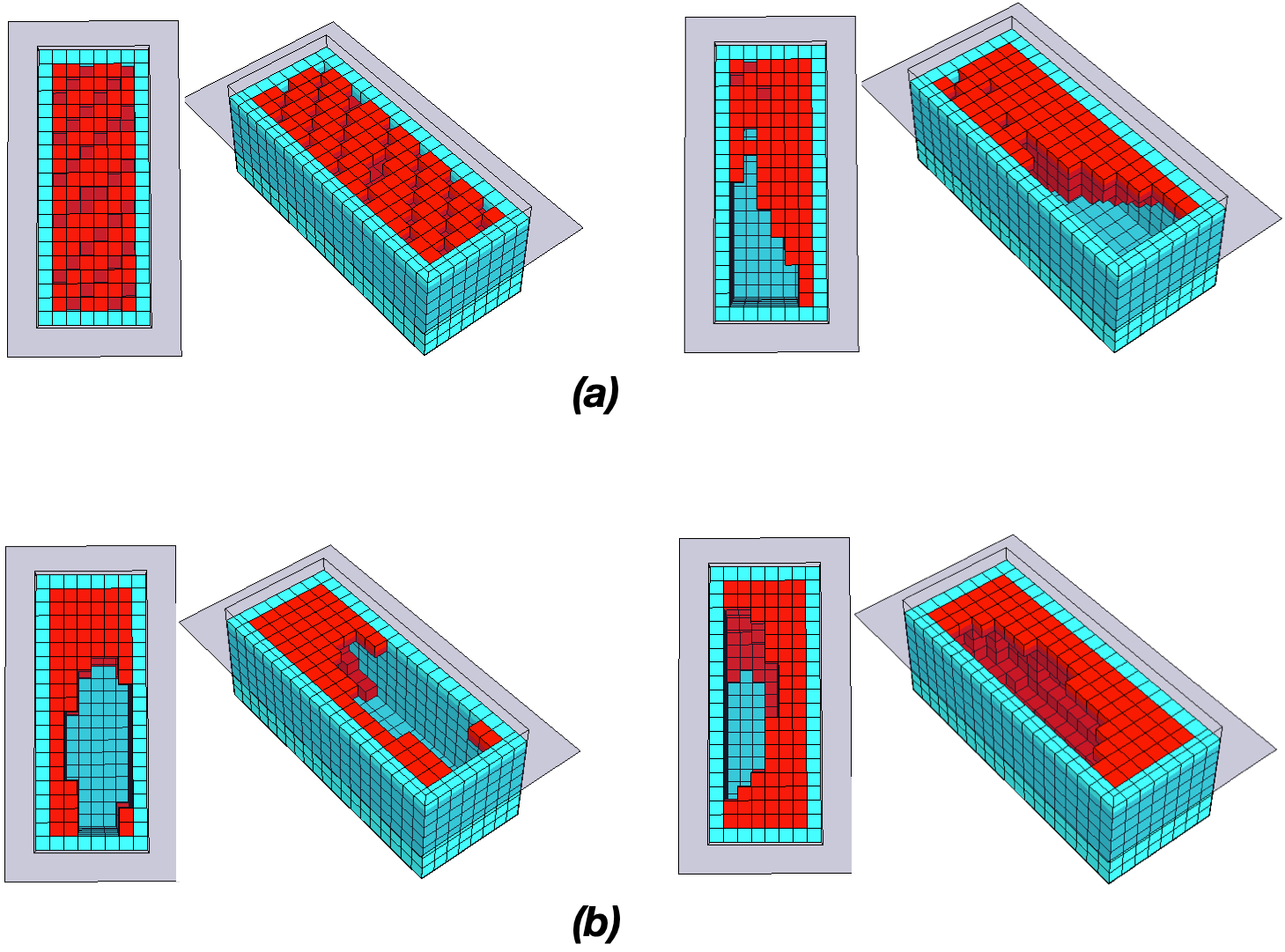}
  \caption{Examples of SAMs found in \cite{Alcaraz2024actuator}: (a) fittest and (b) worst.}
  \label{fig:experimental_evolutionary_configuration_sams}
\end{figure}

Furthermore, all SAMs used in this research have emerged from optimization methods that were described in previous studies \cite{Alcaraz2024actuator}. In specific, two sets of SAMs are utilised: the nine fittest SAMs (NF) (similar to what was reported in \cite{Alcaraz2024actuator} - examples of these are exhibited in Fig.~\ref{fig:experimental_evolutionary_configuration_sams}-a), and the nine worst SAMs (NW) (that are not reported elsewhere - examples of these are exhibited in Fig.~\ref{fig:experimental_evolutionary_configuration_sams}-b). It is noteworthy that while the NF SAMs are expected to have an advantage, since they were the best performing individuals in prior tests, the NW SAMs are also considered here to prove that the evolutionary pressure put towards the proposed methodologies does not affect the robustness of the final produced controlling strategies. The NW SAMs are the worst performing individuals after 3000 generations of optimization methodologies that include niching techniques (for more detail, refer to \cite{Alcaraz2024actuator}), so, they are not considered as totally inept solutions, but they pertain some useful structural functionality.

Finally, due to the significant computational time that simulations require, the implementation of NEAT and HyperNEAT was designed under a client-server architecture to take advantage of distributed computation features. The experimental infrastructure used in this research is replicated from \cite{Alcaraz2024locomotion,Alcaraz2024actuator}. The hardware used is as follows: {\em Processor}: ARM (virtualised), nine cores (18 threads), 3.20 GHz. {\em RAM Memory}: 16 GB, LPDDR5. 


\section{Experimental results}\label{sec:experiments_comparison}

In order to study the suitability of NEAT and HyperNEAT to design controllers for SAMs, four metrics are used: (i) analysing the general performance of controllers in terms of inducing upward bending movements in the $yz$ plane; (ii) evaluating the robustness of controllers; (iii) assessing the complexity of controllers; and (iv) studying the activation functions' influence towards the design process of controllers. Moreover, two different dictionaries of activation functions are used for each metric during experimentation (see Section~\ref{sec:experimental_evolutionary_configuration}). The way each of these metrics are calculated and their purpose are detailed in the following sections.

Furthermore, a standard genetic algorithm (SGA) is utilised as a baseline. The solution representation strategy is based on a bi-dimensional array of real numbers in the $[-2\pi, 2\pi]$ range. The crossover operator implemented is {\em two-point} with probability $0.9$.  For mutation, one matrix element is randomly chosen and replaced by a number in the $[-2\pi, 2\pi]$ range at random. Mutation occurs with a probability of $0.1$. The implementation of the elements of the SGA, such as individuals and the fitness function, is written using {\em O$^3$R}, an object-oriented framework \cite{Alcaraz2022}.

\subsection{General Performance}\label{sec:experiments_comparison_performance}

\begin{figure}[tb!]
  \centering
     \includegraphics[width=1.0\linewidth]{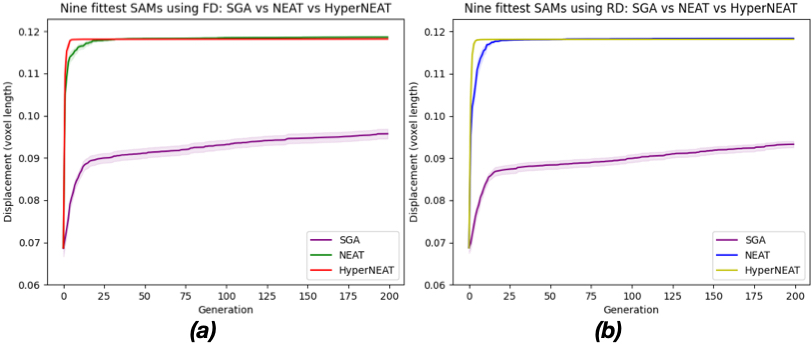}
  \caption{Nine fittest SAMs: Mean general performance with $\pm95\%$ confidence interval (shaded region) under SGA, NEAT, and HyperNEAT using: (a) FD and (b) RD.}
  \label{fig:experiments_comparison_performance_fittest}
\end{figure}

The primary objective of controllers is to induce upward bending movement in the SAMs. This bending movement is measured by the displacement observed in the $yz$ plane of the simulated environment. In this metric, the general performance of SGA, NEAT and HyperNEAT in designing controllers whose main task is inducing upward bending movements to SAMs is studied. Figure~\ref{fig:experiments_comparison_performance_fittest} shows the mean performance of the fittest controller in each population with $\pm95\%$ confidence interval exhibited by the shaded region throughout 20 duplicate evolutionary runs under SGA, NEAT and HyperNEAT using NF for both variations of the activation function dictionaries.

When FD is utilised (see Fig.~\ref{fig:experiments_comparison_performance_fittest}-a), NEAT and HyperNEAT can find the fittest controller within the first few generations of the evolutionary process. In contrast, SGA exhibits a slower evolutionary pace. The performance of the three approaches presents significant differences despite the fact that the difference between NEAT and HyperNEAT performances is not visible within the $y$-axis used to illustrate all three methods. The data collected were tested, and it is not normally distributed (Shapiro-Wilk test; $p<0.01$). Then, the Kruskal-Wallis was used to confirm significant differences among the performances of SGA, NEAT and, HyperNEAT ($p<0.01$). Thus, it is possible to rank the performance of the three algorithms: NEAT $>$ HyperNEAT $>$ SGA (Dunn's test $p<0.01$).

Furthermore, when RD is implemented (see Fig.~\ref{fig:experiments_comparison_performance_fittest}-b), again, NEAT and HyperNEAT find the fittest controller at the early stages of evolution, whereas SGA tends to evolve slower. Significant differences exist among the performances of the three approaches. The data gathered were tested and proved not normally distributed (Shapiro-Wilk test; $p<0.01$). It is feasible to confirm significant differences among the performance of the three approaches by employing the Kruskal-Wallis test ($p<0.01$). Therefore, a performance ranking of the performances can be performed: NEAT $>$ HyperNEAT $>$ SGA (Dunn's test; $p<0.01$).

\begin{figure}[tb!]
  \centering
     \includegraphics[width=1.0\linewidth]{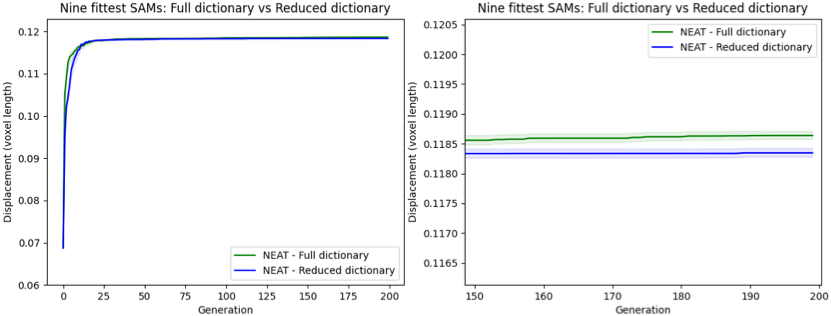}
  \caption{Nine fittest SAMs: Left - Mean general performance with $\pm95\%$ confidence interval (shaded region) under NEAT using FD (green curve) and RD (blue curve); Right - Close up of the mean general performance of NEAT in the last 50 generations using FD (green curve) and RD (blue curve).}
  \label{fig:experiments_comparison_performance_fittest_best}
\end{figure}

In addition, Fig.~\ref{fig:experiments_comparison_performance_fittest_best} presents the mean performance of the fittest controller with $\pm95\%$ confidence interval exhibited by the shaded region throughout 20 evolutionary runs under NEAT employing FD and RD. The performance induced by the two dictionaries presents significant differences (paired t-test; Wilcoxon test:  $p<0.01$). In general, NEAT can find more suitable controllers when FD is implemented due to the higher number of activation functions available and, thus, the ability to administer a more complicated behaviour. 
However, the difference in displacement is quite small, which may not justify the inherent overheads of a broader search space.

\begin{figure}[tb!]
  \centering
     \includegraphics[width=1.0\linewidth]{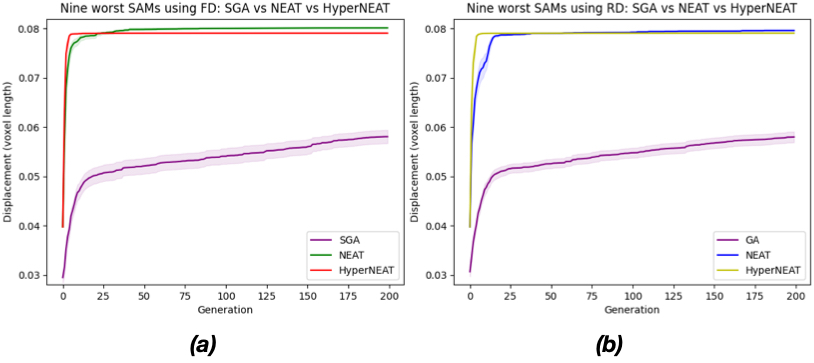}
  \caption{Nine worst SAMs: Mean general performance with $\pm95\%$ confidence interval (shaded region) under SGA, NEAT, and HyperNEAT using: (a) FD and (b) RD.}
  \label{fig:experiments_comparison_performance_worst}
\end{figure}

So far, the experiments performed used NF, which can be considered SAMs whose morphology can be controlled with ease. The following experiments use NW, a set of SAMs that can be deemed significantly more challenging to control than NF. Figure~\ref{fig:experiments_comparison_performance_worst} presents the mean performance of the fittest controller per generation with $\pm95\%$ confidence interval shown by the shaded region throughout 20 duplicate evolutionary runs under SGA, NEAT and HyperNEAT using NW.

When FD is implemented (see Fig.~\ref{fig:experiments_comparison_performance_worst}-a), NEAT and HyperNEAT find the fittest controller during the first generations. On the other hand, SGA presents a slower evolutionary pace. Significant differences exist among the performance of the three algorithms. The data collected were tested and presented a non-normal distribution (Shapiro-Wilk test; $p<0.01$). Then, confirming significant differences by the Kruskal-Wallis test is possible ($p<0.01$). Consequently, a performance rank can be conducted: NEAT $>$ HyperNEAT $>$ SGA (Dunn's test; $p<0.01$).

Moreover, when RD is used (see Fig.~\ref{fig:experiments_comparison_performance_worst}-b), again, the NE-based approaches discover the fittest controller during the first generations of the evolutionary process. Among the performances of the three approaches, significant differences can be observed (Shapiro-Wilk test; Kruskal-Wallis test: $p<0.01$). Thus, their performances can be ranked as follows: NEAT $>$ HyperNEAT $>$ SGA (Dunn's test; $p<0.01$).

\begin{figure}[tb!]
  \centering
     \includegraphics[width=1.0\linewidth]{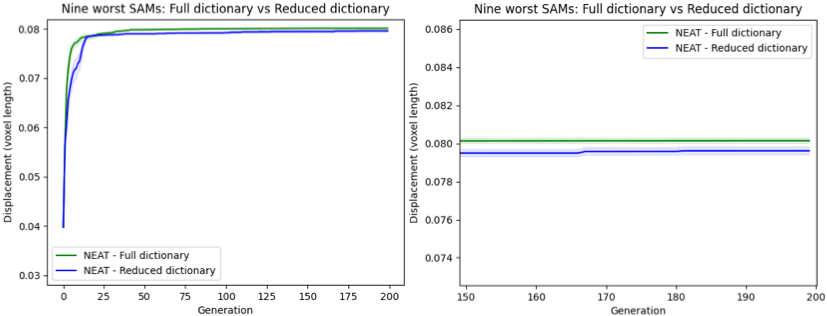}
  \caption{Nine worst SAMs: Left - Mean general performance with $\pm95\%$ confidence interval (shaded region) under NEAT using FD (green curve) and RD (blue curve); Right - Close up of the mean general performance of NEAT in the last 50 generations using FD (green curve) and RD (blue curve).}
  \label{fig:experiments_comparison_performance_worst_best}
\end{figure}

Figure~\ref{fig:experiments_comparison_performance_worst_best} shows the mean performance of the fittest controller with $\pm95\%$ confidence interval exhibited by the shaded region over 20 evolutionary runs under NEAT implementing FD and RD. The performance of NEAT using the two different dictionaries exhibits significant differences (paired t-test; Wilcoxon test: $p<0.01$). As shown, NEAT can discover fitter individuals (i.e., controllers) when FD is used. This outcome arguably occurs because FD is composed of more activation functions than RD, which implies a broader exploration of the search space and, therefore, the possibility of discovering fitter individuals. Although, the difference in displacement is minimal.

The results suggest that NEAT and HyperNEAT can find fitter controllers than those found by SGA in fewer generations due to their capacity to consider two aspects of SAMs during the evolutionary process: (i) the morphology, and (ii) the type of material. 
Finally, NEAT exhibited the best performance regardless of the dictionary employed due to its mechanism allowing evolution to explore a broader search space where the number of connections and the number of neurons are not limited.

\subsection{Robustness}\label{sec:experiments_comparison_robustness}

A controller may exhibit suitable performance in a certain scenario (i.e., a specific SAM under specific conditions of the environment). Nevertheless, the performance may be different when just one aspect of the scenario changes. This metric focuses on analysing the suitability of controllers to induce upward bending movements regardless if the SAM changes. The robustness of controllers (i.e., aptitude) is computed utilising the displacement observed in the $yz$ plane of nine different SAMs. Equation~\ref{eq:experiments_comparison_robustness} is used to calculate the aptitude induced by the evolved controllers.

\begin{equation}\label{eq:experiments_comparison_robustness} 
    apt_h= \frac{\sum_{i=1}^{9} displacement_i}{9}
\end{equation}

\noindent
where $apt_h$ is the aptitude of controller $h$, and $displacement_i$ is the displacement observed of the $i$-th SAM composing a group of SAMs (see Section~\ref{sec:experimental_evolutionary_configuration}).

\begin{figure}[tb!]
  \centering
     \includegraphics[width=1.0\linewidth]{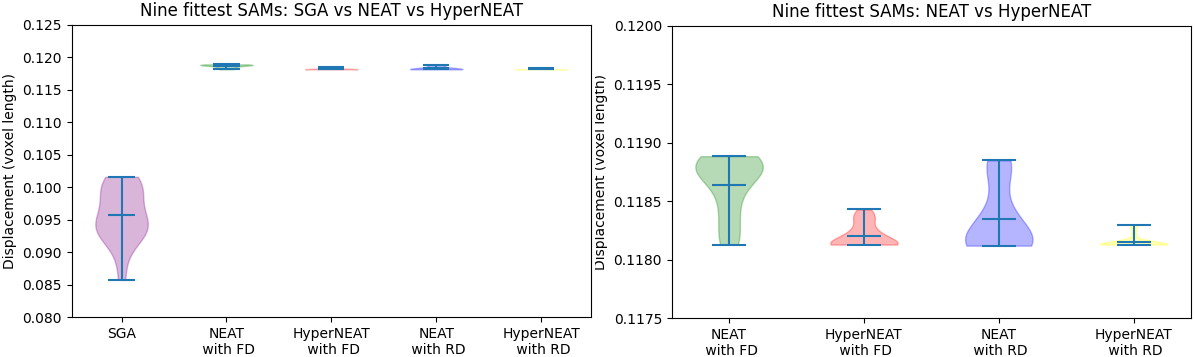}
  \caption{Nine fittest SAMs: Left - Displacement observed in the $yz$ plane under SGA, NEAT, and HyperNEAT, using FD and RD; Right - Close up of the displacement observed in the $yz$ plane under NEAT and HyperNEAT, using FD and RD.}
  \label{fig:experiments_comparison_robustness_fittest}
\end{figure}

Figure~\ref{fig:experiments_comparison_robustness_fittest} shows violin plots that compare the displacement induced by the fittest individual (i.e., controller) discovered by SGA, NEAT and HyperNEAT using FD, RD, and the nine SAMs utilising NF across 20 evolutionary trials. Each violin plot exhibits minimum, maximum, median, and kernel density estimation of the frequency distribution of the values.

The displacement induced by all the approaches present some significant differences. All the data gathered were tested and are not normally distributed (Shapiro-Wilk test; $p<0.01$). Then, using the Dunn's test, it is possible to confirm that there are not significant differences among HyperNEAT with FD, NEAT with RD, and HyperNEAT RD ($p>0.05$). In contrast, there are significant differences between: (a) all the NE-based approaches and SGA, and (ii) NEAT using FD and the rest of the NE-based approaches ($p<0.01$). Thus, a robustness rank can be performed: NEAT using FD $>$ HyperNEAT with FD, NEAT with RD, HyperNEAT with RD $>$ SGA.

\begin{figure}[tb!]
  \centering
     \includegraphics[width=1.0\linewidth]{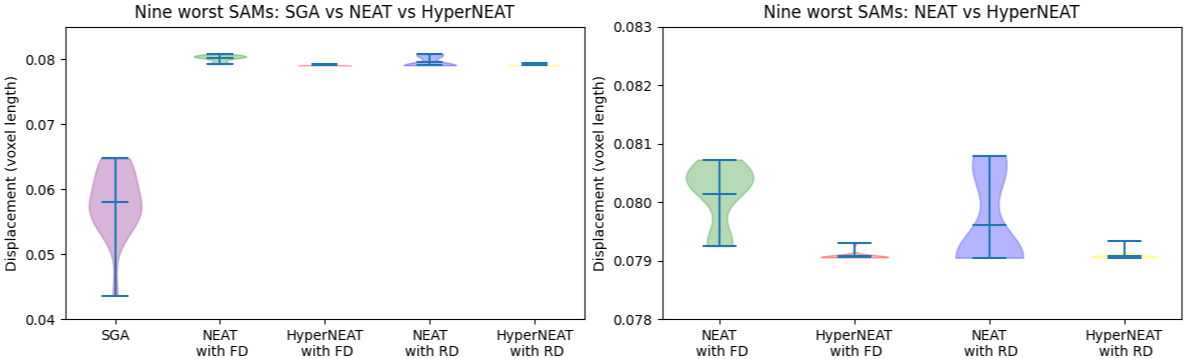}
  \caption{Nine worst SAMs: Left - Displacement observed in the $yz$ plane under SGA, NEAT, and HyperNEAT, using FD and RD; Right - Close up of the displacement observed in the $yz$ plane under NEAT and HyperNEAT, using FD and RD.}
  \label{fig:experiments_comparison_robustness_worst}
\end{figure}

Furthermore, Fig.~\ref{fig:experiments_comparison_robustness_worst} depicts violin plots comparing the displacement provoked by the fittest controller found by SGA, NEAT and HyperNEAT using FD, RD, and the nine SAMs of NW across 20 evolutionary runs. Each violin plot shows the median, maximum, minimum, and  kernel density estimation of the frequency distribution of the values. 

Some significant differences can be observed in the displacement of all algorithms. The data collected during experimentation were tested and presented a non-normal distribution (Shapiro-Will test; $p<0.01$). Through the Dunn's test, it is feasible to confirm, again, that not significant differences among HyperNEAT with FD, NEAT with RD, and HyperNEAT with RD exist ($p>0.05$). On the other hand, significant differences can be observed between NEAT using FD and the rest of the NE-based approaches, as well as all the NE-based approaches and SGA ($p<0.01$). Based on this analysis, it is possible to rank the robustness of the algorithms: NEAT using FD $>$ HyperNEAT with FD, NEAT with RD, HyperNEAT with RD $>$ SGA.

The results point out that NEAT and HyperNEAT can discover more suitable controllers for inducing higher upward bending movements than those found by SGA regardless of: (i) the number of SAMs implied during evolution, and (ii) the morphological complexity of soft actuators. The outperforming behaviour exhibited by the NE-based approaches arguably is due to CPPNs, which are an essential element of their mechanism, allowing the generation of patterns that help to enhance the efficiency of movements induced in SAMs. Finally, NEAT consistently shows the ability to find controllers with the best robustness regardless of the dictionary used since, during evolution, it explores the search space where the number of neurons and, hence, the number of connections are not restricted.

\subsection{Controller Complexity}\label{sec:experiments_comparison_complexity}

Taking into account that SAMs should represent devices that can be built in real life \cite{Alcaraz2024actuator}, the controllers associated with them and designed by the proposed methods must be suitable for real life implementation. Thus, a controller with a simple topology (i.e., fewer nodes and connections) is considered a better and more efficient solution.

In this metric, SGA is not considered due to the poor performance shown in previous experiments (see Section~\ref{sec:experiments_comparison_performance} and Section~\ref{sec:experiments_comparison_robustness}). Moreover, it is indispensable to consider that under NEAT, controllers are represented by CPPNs. On the other hand, under HyperNEAT, substrates (see Section~\ref{sec:experimental_setup_configuration_hyperneat}) represent controllers. Table~\ref{tab:experiments_comparison_complexity} presents the mean number of connections and hidden neurons (i.e., nodes) composing the fittest controller under NEAT and HyperNEAT throughout 20 evolutionary runs, using the two activation function dictionaries variants and the two SAM groups. 

\begin{table}
\caption{Mean number of connections and hidden neurons of the fittest controller designed by NEAT and HyperNEAT using activation function dictionaries and SAM groups.}
\label{tab:experiments_comparison_complexity}
 \begin{center}
  \begin{tabular}{ c c c c c } 
   \toprule
   Approach & SAM Group & Dictionary & Connections & Hidden nodes  \\
   \midrule
    NEAT & NF & FD & 2 & 3 \\
    NEAT & NF & RD & 2 & 2 \\ 
    NEAT & NW & FD & 3 & 3 \\
    NEAT & NW & RD & 3 & 3 \\
    HyperNEAT & NF & FD & 36 & 13 \\
    HyperNEAT & NF & RD & 39 & 13 \\ 
    HyperNEAT & NW & FD & 43 & 13 \\
    HyperNEAT & NW & RD & 33 & 13 \\
   \bottomrule
  \end{tabular}
 \end{center}
\end{table}

The results indicate that, regardless of the activation function dictionary variant and the morphological complexity of soft actuators, HyperNEAT generates more complex controllers than those generated by NEAT, due to the number of hidden neurons composing the substrate being fixed (see Fig.~\ref{fig:experimental_setup_configuration_hyperneat_substrate}). This limitation, in terms of hidden neurons, narrows the exploration of the search space since the number of connections, their weights, and the bias of neurons need to be optimised. In contrast, NEAT can explore a broader search space that considers the number of neurons and their bias, the number of connections and consequently their weights, and the activation functions within neurons. These features allow NEAT to discover more efficient topologies and, consequently, more suitable controllers that can be built in real life than those found by HyperNEAT.

\subsection{Explainability}\label{sec:exp_comparison_explainability}

Thus far, the results obtained from previous metrics have shown that NEAT is the most suitable approach for finding adequate controllers for SAMs. Furthermore, it is feasible to conclude that CPPNs, the primary mechanism of NEAT, are a significant element that helps to enhance the upward bending movements induced by controllers to SAMs. However, the previous metrics did not investigate in detail how CPPNs operated to achieve the results obtained. Therefore, this metric, which only considers NEAT due to its higher performance, is focused on analysing the role of the activation functions utilised in the neurons within the topology of the fittest individuals found during evolution. 

\begin{figure}[tb!]
  \centering
     \includegraphics[width=1.0\linewidth]{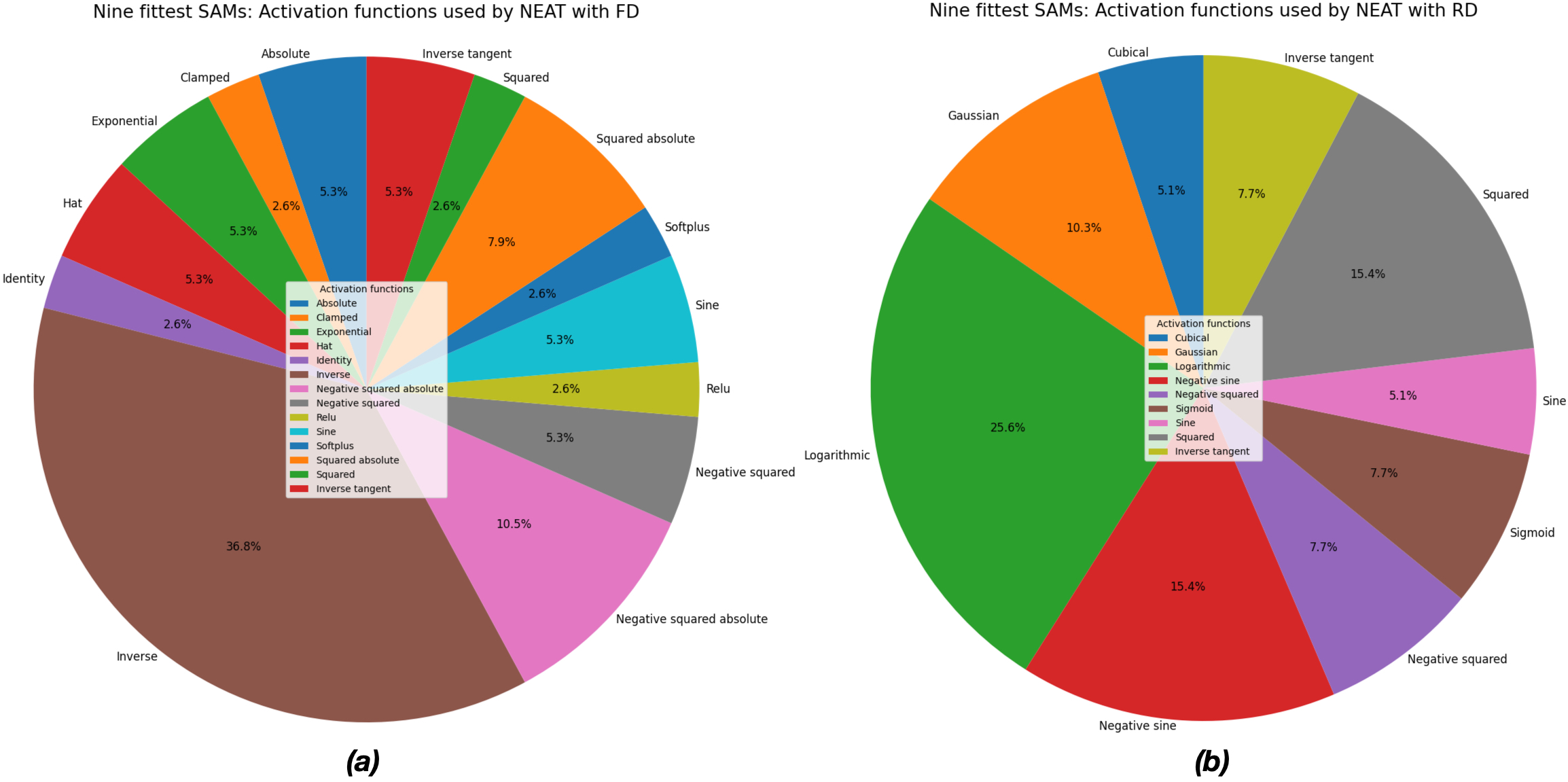}
  \caption{Nine fittest SAMs: Activation functions employed to compose the fittest controllers found by NEAT using: (a) FD, and (b) RD.}
  \label{fig:experiments_comparison_explainability_fittest}
\end{figure}

Figure~\ref{fig:experiments_comparison_explainability_fittest} presents pie plots showing the activation functions utilised to compose the fittest controllers using NF. Each pie plot exhibits the usage, in terms of percentage, of each activation function implied.

When FD is used (see Fig.~\ref{fig:experiments_comparison_explainability_fittest}-a), the activation functions most used are {\em Inverse} with 36.8\%, followed by {\em Negative squared absolute} with 10.5\% and {\em Squared absolute} with 7.9\%. NEAT utilised the sinusoidal-like functions and their variations available in FD to generate a periodic pattern of the output of controllers. Furthermore, the {\em Inverse} function is highly used because it arguably helps to complete an entire period of the output signal of controllers. Moreover, when RD is employed (see Fig. \ref{fig:experiments_comparison_explainability_fittest}-b), the most used activation functions are {\em Logarithmic} with 25.6\%, followed by {\em Negative sine} and {\em Squared} with 15.4\%. Again, NEAT used those activation functions that generate periodic patterns in the output of controllers. The {\em Logarithm} function is widely used by NEAT, arguably due to the signal needing a continuous and steady increment when controllers process it. 

In general, due to the geometric patterns observed in SAMs composing NF, such as a striped diagonal pattern with no voxels throughout the contractile body composing the SAM (see Fig.~\ref{fig:experimental_evolutionary_configuration_sams}; left) and a pyramidal-like pattern having a few contractile voxels at the bottom and gradually incrementing the number of voxels towards the top of the actuator (see Fig.~\ref{fig:experimental_evolutionary_configuration_sams}; right), a periodic signal (i.e., sinusoidal-like) is more convenient to maximise the upward bending movements induced by controllers to SAMs. 

\begin{figure}[tb!]
  \centering
     \includegraphics[width=1.0\linewidth]{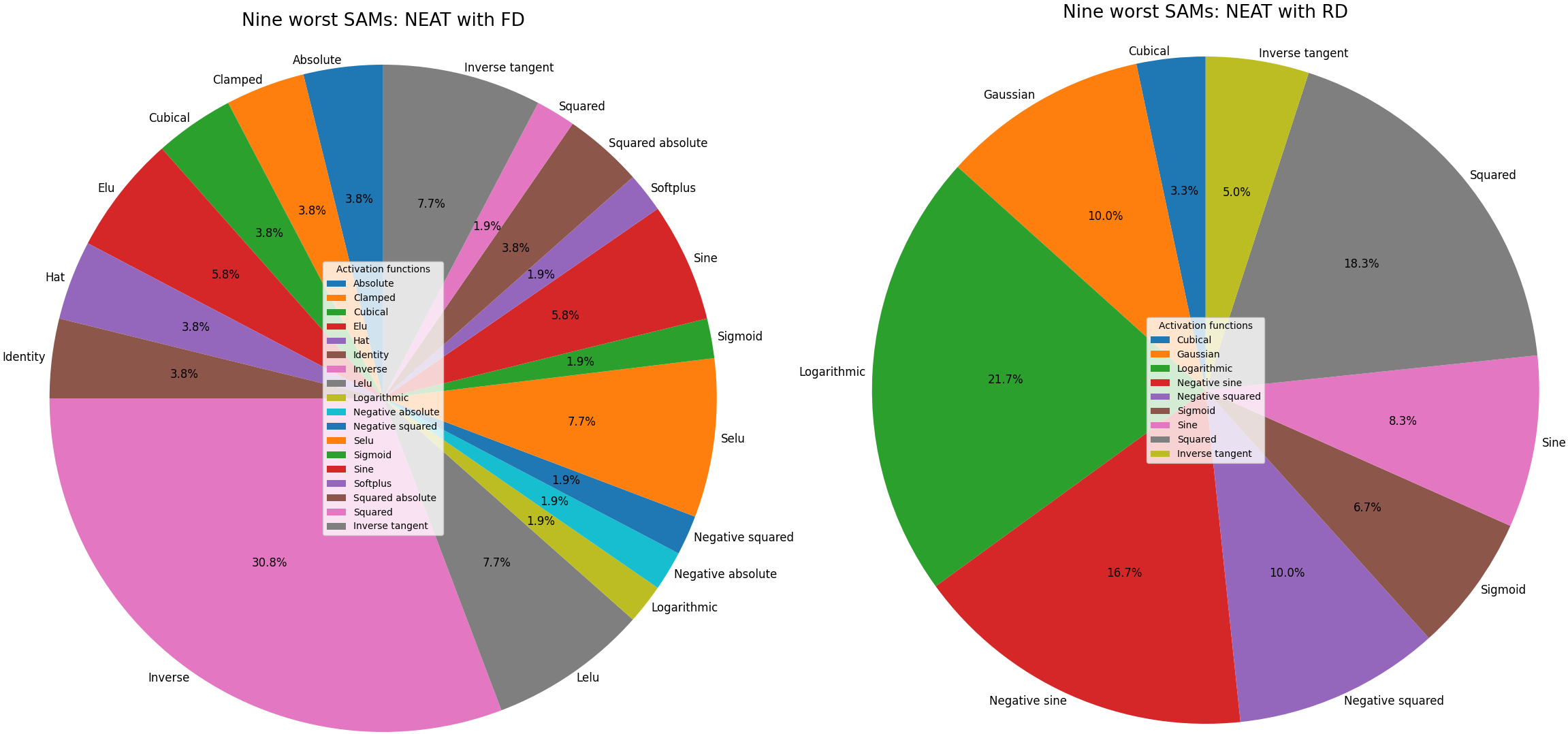}
  \caption{Nine worst SAMs: Activation functions employed to compose the fittest controllers found by NEAT using: (a) FD, and (b) RD.}
  \label{fig:experiments_comparison_explainability_worst}
\end{figure}

Figure~\ref{fig:experiments_comparison_explainability_worst} depicts pie plots presenting the activation functions utilised to compose the fittest controllers using NW. The pie plots exhibit the usage, in terms of percentage, of each activation function implied.

For the case where FD is implemented (see Fig.~\ref{fig:experiments_comparison_explainability_worst}-a), the following functions are predominantly used: {\em Inverse} function with 30.8\% followed by {\em Lelu}, {\em Selu} and {\em Inverse tangent} with 7.7\%. NEAT employs functions that trigger an abrupt increment in the control signal. In addition, similar to the case when NF is used, the {\em Inverse} function is used several times, arguably because of its usefulness to generate an entire period of the output signal. Furthermore, when RD is implemented (see Fig.~\ref{fig:experiments_comparison_explainability_worst}-b), the most used activation functions are {\em Logarithmic} with 21.7\%, followed by {\em Squared} with 18.3\%, {\em Negative sine} with 16.7\% and {\em Negative squared} with 10\%. In this scenario, NEAT employed periodic functions that induce a periodic pattern in the output of controllers. Regarding the {\em Logarithm} function, and in a similar fashion to the scenario when NW is used, it is employed by NEAT to provide a continuous and steady increment to the signal when controllers are processing it. 

Due to the lack of symmetrical patterns observed in SAMs composing NW (see Fig.~\ref{fig:experimental_evolutionary_configuration_sams}-b), for instance, a significant voxel discontinuity in the contractile body composing SAMs, it is necessary for NEAT to use a function that significantly rises the output signal (i.e., a peak-like) of controllers to induce as much as possible the upward bending movement to SAMs.


\section{Conclusions}\label{sec:conclusions}

This research 
assesses the capacity of NEAT and HyperNEAT to design suitable controllers for SAMs that represent a catheter for targeted drug delivery. An SGA is utilised as a baseline. The design properties of the three approaches are evaluated by four metrics: (i) analysing the general performance in terms of reaching the maximum upward bending movement to SAMs; (ii) evaluating the robustness of controllers generated; (iii) studying the complexity in terms of number of nodes and their connections of controllers designed; and (iv) investigating the effect of the activation functions in the design of controllers. In addition, two different activation function dictionaries and two sets of SAMs are utilised for experimentation for the four metrics (a full dictionary -as in previous works \cite{Alcaraz2024controllers}- and a reduced one). It is necessary to highlight that metrics (iii) and (iv) are only applied to controllers designed by the NE-based algorithms.

The provided results suggest that NEAT and HyperNEAT are more suitable for designing controllers than traditional approaches (i.e., a standard genetic algorithm) despite of: (i) the morphological complexity of the soft actuators; (ii) the number of activation functions available during evolution; and (iii) the number of SAMs used as testbeds during the evolutionary process. In general, NEAT and HyperNEAT outperform SGA due to: (a) their ability to embody the morphology and the materials of SAM during the design (i.e., evolutionary) process of controllers, and (b) having  CPPNs as the core elements, whose inherent topological features facilitate the generation of better control patterns that substantially enhance the efficiency of upward bending movements induced by controllers to SAMs.

When the performance of NEAT and HyperNEAT is compared, although the differences observed are minimal, they are statistically significant. Thus, it is concluded that NEAT performs better, since the search space where it operates considers the number of neurons, their connections and activation functions. In contrast, HyperNEAT works in a more limited search space because only the connections of a fixed number of neurons are considered. Regarding the unexpected behaviour of HyperNEAT, two aspects arguably affected its performance: (i) no geometrical aspects of the domain problem are available that can be embodied in the design of the substrate used, and (ii) the design of the substrate used for experimentation was implemented only considering the allocation of the neurons in a two-dimension space.

Another interesting insight of this research is the relationship between the domain problem (i.e., the SAMs) and the activation functions used by NEAT and HyperNEAT. Both algorithms use sinusoidal-like functions when the contractile bodies of SAMs exhibit patterns that are either symmetric or geometrical. On the other hand, when the patterns are not symmetrical or geometric, NEAT and HyperNEAT utilise activation functions that present a peak-like response.  

Considering the results obtained in this research, future work directions have been outlined in the following. For instance,  to test the reliability of NEAT and HyperNEAT, simulations of SAMs in an environment where other elements, such as friction and viscosity, may be included. Furthermore, exploring different sets of activation functions to analyse the performance of the NE-based is also considered. In addition, improving the performance of HyperNEAT with a broader exploration of the number of neurons, hidden layers, and activation functions used in the substrate is another avenue for future work. Finally, other NE-based approaches can be used to perform a more robust benchmark, for instance, an extension of HyperNEAT named {\em ES-HyperNEAT} \cite{Risi2012} and Evolutionary eXploration of Augmenting Memory Models (EXAMM) \cite{Ororbia2019, Thakur2023}.

\section*{Acknowledgement}
This project has received funding from the European Union’s Horizon Europe research and innovation programme under grant agreement No. 101070328. UWE researchers were funded by the UK Research and Innovation grant No. 10044516.

\end{document}